\pdfoutput=1

\documentclass[11pt]{article}

\usepackage[final]{mtsummit25}
 \usepackage{copyright}

\usepackage{times}
\usepackage{latexsym}
\usepackage{subfig} 

\usepackage[T1]{fontenc}

\usepackage[utf8]{inputenc}
\usepackage{colortbl} 
\usepackage{booktabs}

\usepackage{microtype}

\usepackage{inconsolata}

\usepackage{graphicx}

\title{Do Not Change Me: On Transferring Entities Without Modification in Neural Machine Translation – a Multilingual Perspective}

\author{
  \textbf{Dawid Wisniewski\textsuperscript{1,2}},
  \textbf{Mikolaj Pokrywka\textsuperscript{1,3}},
  \textbf{Zofia Rostek\textsuperscript{1}}
\\
  \textsuperscript{1}Laniqo.com,
  \textsuperscript{2}Poznan University of Technology, Poland
  \textsuperscript{3}Adam Mickiewicz University, Poland,
\\
  \small{
    \textbf{Correspondence:} \href{mailto:email@domain}{dawid.wisniewski@laniqo.com}
  }
}

\begin{document}
\maketitle
\begin{abstract}
Current machine translation models provide us with high-quality outputs in most scenarios. However, they still face some specific problems, such as detecting which entities should not be changed during translation. In this paper, we explore the abilities of popular NMT models, including models from the OPUS project, Google Translate, MADLAD, and EuroLLM, to preserve entities such as URL addresses, IBAN numbers, or emails when producing translations between four languages: English, German, Polish, and Ukrainian. We investigate the quality of popular NMT models in terms of accuracy, discuss errors made by the models, and examine the reasons for errors. Our analysis highlights specific categories, such as emojis, that pose significant challenges for many models considered. In addition to the analysis, we propose a new multilingual synthetic dataset of 36,000 sentences that can help assess the quality of entity transfer across nine categories and four aforementioned languages\footnote{Project supported by grant no. 0311/SBAD/0763 - Mloda Kadra financed by Poznan University of Technology.
}.

\end{abstract}

\section{Introduction}

Machine translation is one of the oldest branches of Natural Language Processing (NLP), which, thanks to the Transformer~\cite{DBLP:conf/nips/VaswaniSPUJGKP17} architecture incorporating the (self-)attention~\cite{DBLP:journals/corr/BahdanauCB14} mechanism, achieves human-like quality in many translation directions, especially in the case of sentence-level translations~\cite{DBLP:conf/emnlp/LaubliS018}.

Although the overall quality of recent neural models is very high both in terms of human perception and metrics such as BLEU~\cite{DBLP:conf/acl/PapineniRWZ02} or COMET~\cite{DBLP:conf/wmt/ReiSAZFGLCM22}, there are still some weaknesses that need to be addressed.

One of them is the problem of identifying text fragments that should be copied without modification into the target sentence. Many of these entities: phone numbers, email addresses, or company names, represent categories that occur frequently in texts. However, the possible instantiations of these categories are so numerous (e.g., all possible phone numbers or email addresses) that models must rely on contextual information to detect them, rather than memorizing these entities.

In this paper, we show that many popular NMT models have problems with transferring such entities without modification. To analyze this issue, we focus on translations between four languages: English, German, Polish, and Ukrainian, using sentences containing entities from 9 categories: alphanumeric sequences, emails, emojis, IBANs, IP numbers, ISBNs, phone numbers, social handlers, and URLs.

The main contributions of this paper are: (i) a new multilingual dataset consisting of 36,000 sentences, each of which contains entities that should not be modified in the translation process, (ii) an analysis of eight popular NMT models, and (iii) a discussion of the causes of errors made by the models considered. This paper addresses three research questions: \textbf{RQ1}: Are there categories of entities that popular NMT models cannot transfer without modification? \textbf{RQ2}: What are the causes of the problems with transferring entities without modification? \textbf{RQ3}: Which models offer the highest quality solutions to this problem?


\section{Related work}
\paragraph{Translation errors} 
In recent years, much attention has been paid to the evaluation of neural machine translation (NMT) models. Although the Transformer-era NMT models generally perform very well, some of them are prone to small but sometimes critical errors. The problem analyzed in this paper has so far been considered in the context of the sensitivity of evaluation metrics to critical errors, among which unexpected input sequence modification has often been selected as one of the critical error types.

ACES~\cite{amrhein-etal-2022-aces} -- the set of translation accuracy challenges, used to assess the sensitivity of metrics to critical errors, is a data set consisting of approximately 36,500 translated sentences. These sentences are expressed in 146 languages and describe 68 error phenomena, including hallucinations, erroneous unit conversions, adding unnecessary information, producing nonsense words, or translating entities that should not be translated. The error categories in ACES follow the Multidimensional Quality Metrics (MQM) ontology, providing a taxonomy of 108 translation problems defined at multiple levels of granularity~\cite{lommel2014multidimensional}. 

Another similar dataset, DEMETR~\cite{demetr-2022}, was developed to analyze machine translation metrics. This dataset contains examples of 35 perturbations that include: deleting parts of speech, using hyperonyms, replacing words, misspelling, using wrong capitalization, or repeating parts of text. 

SMAUG~\cite{alves-etal-2022-robust}, a sentence-level multilingual augmentation project focuses on generating translations that include critical errors and may be used to evaluate the robustness of MT metrics. SMAUG is a tool for introducing perturbations in existing sentences and is focused on categories such as deviation from named entities, numbers or meaning, or modification of content. 

However, our goal is to analyze the problem from a different perspective. Instead of perturbing existing translations, we observe which models are more likely to generate perturbations. Furthermore, we increase the number of examples per category, which helps us to draw statistically significant conclusions. In our work, we provide 1,000 examples per category and language pair, while in other works, e.g., ACES, 36,500 examples cover 146 languages and 68 phenomena, which gives an average of about 3.5 examples per language and category combination. Moreover, instead of focusing only on translations from a given language to English, as in the DEMETR dataset, we analyze all possible translation directions between the four languages under consideration.

\paragraph{Recent neural translation models}

Since the advent of the Transformer~\cite{DBLP:conf/nips/VaswaniSPUJGKP17} architecture, the quality of machine translation models has increased by a large margin~\cite{DBLP:journals/jair/Stahlberg20}. For this reason, multiple Transformer-based MT models have been proposed in recent years. One of the well-known projects in this area is OPUS~\cite{DBLP:journals/lre/TiedemannABBGNRSVV24}, which provides datasets and MT models trained using MARIAN~\cite{mariannmt} and specialized in translation between various pairs of languages. Other similar models try to explore the multilingual abilities of Transformers, e.g., MBART~\cite{DBLP:journals/corr/abs-2008-00401} supporting translation in 50 languages, No Language Left Behind (NLLB)~\cite{DBLP:journals/corr/abs-2207-04672} supporting 200 languages, or MADLAD~\cite{DBLP:conf/nips/KuduguntaC0GXKS23} with the support of more than 400 languages. These models are trained using multilingual corpora, which help to exploit relationships between languages.
In addition to scaling models in terms of language number, other approaches try to add different modalities as additional sources of knowledge. A popular example is SeamlessM4T~\cite{DBLP:journals/corr/abs-2308-11596}, which is able to process textual and audio data.

An interesting avenue in machine translation is the use Large Language Models (LLMs) to translate between languages~\cite{DBLP:conf/wmt/WuH23}. Due to the multilingualism of large web corpora and the variety of tasks observed in these corpora that can help to better understand the language, these general models are an interesting alternative to specialized translation models~\cite{DBLP:journals/corr/abs-2310-06825,DBLP:journals/corr/abs-2407-21783,DBLP:journals/corr/abs-2408-00118,DBLP:journals/corr/abs-2407-10759}. Some LLMs are trained with a substantial amount of parallel corpora, for example, TowerLLM~\cite{DBLP:journals/corr/abs-2402-17733}, which supports 10 languages and, in addition to machine translation, can perform grammatical correction of texts, identify named entities, or post-edit texts. EuroLLM~\cite{DBLP:journals/corr/abs-2409-16235} is another attempt to use LLMs for machine translation. This model supports 35 languages and is trained on various sources of data, including good-quality parallel translation corpora, mathematical equations, code, and general web data. 

\section{Dataset}
The ACES, DEMETR, and SMAUG have different scopes than this research; thus, even if they consider entities that should not be translated (hereafter referred to as \textit{no-translate entities}), they use them to evaluate metrics. They also provide relatively small numbers of examples regarding no-translate entities. For example, ACES introduces only 100 examples representing the no-translate category, all of them representing English to German translations. DEMETR focuses on translating from one of 10 languages to English, providing 100 examples per a given perturbation and source language pair. SMAUG operates on existing sentences and can be used to introduce perturbations (e.g., punctuation removal, random sequence injection, or named entity modification) into provided sentences.

For this reason, we decided to create a new dataset, entirely focused on entities that should not be modified during translation. This dataset can be used to measure how well different models transfer no-translate entities without modification.

We selected 9 entity types that are simple to identify using regular expressions, these are: \textit{e-mail addresses}, \textit{URLs}, \textit{phone numbers}, \textit{emojis},  \textit{social handlers} (e.g.,  Instagram or TikTok user identifiers), \textit{IP addresses}, \textit{alphanumeric sequences} -- artificial identifiers represented as mixtures of letters and numbers, International Bank Account Numbers (\textit{IBANs}), and International Standard Book Numbers (\textit{ISBNs}). We did not include dates and numbers as their formats may vary depending on the language (e.g., DD/MM/YYYY vs. MM/DD/YYYY dates or 1,000.00 vs. 1.000,00 as representations of "one thousand" in different languages). 

In addition to analyzing individual categories, we wanted to investigate the transferability of entities across languages. We selected 4 languages: English, German, Polish, and Ukrainian — with the goal of selecting popular languages (English, German) and less popular ones that have interesting features, e.g. strongly inflected languages (Polish) or non-Latin alphabets (Ukrainian with Cyrillic). These 9 categories and 4 languages were used for the subsequent generation of the dataset.

\subsection{Sentences generation}
We asked Gemma 2~\cite{DBLP:journals/corr/abs-2408-00118} 9B instruction-following model~\footnote{\url{https://huggingface.co/google/gemma-2-9b-it}} to generate 20,000 sentences expressed in each of the languages considered (English, German, Polish, Ukrainian) and provide examples for each category considered. The prompts used to generate the examples are listed in Appendix~\ref{appendix:generation}. 

To generate sentences, we used vLLM~\cite{DBLP:conf/sosp/KwonLZ0ZY0ZS23} library setting the following sampling parameters: \texttt{temperature=1.2}, \texttt{top\_p=0.95}, \texttt{length\_penalty=1.0}, \texttt{min\_tokens=16}, \texttt{max\_tokens=512},
\texttt{max\_model\_len=4096}, \texttt{gpu\_memory\_utilization=0.95}.

The result of this step was a set of 720,000 generated sentences (20,000 sentences $\times$ 4 languages $\times$ 9 categories).

\subsection{Representative sample selection}
\label{sample_selection}
As LLM outputs may be of varying quality, we performed a sampling procedure to retain only the most representative sentences. 

We applied the following procedure for this purpose: (i) we discarded additional information occasionally generated by Gemma, which frequently consists of the English translation of the sentence, comments about entities, etc. These additional texts were added mainly due to the length constraint \texttt{min\_tokens=16}, so that for short sentences generated, the model continued the generation process to meet the criterion (e.g., providing English translation for languages other than English). Gemma frequently places these additional remarks after double newlines so they can be easily removed, (ii) we discarded examples expressed in a language other than the expected one using the  \texttt{langdetect}~\footnote{\url{https://pypi.org/project/langdetect/}} library, (iii) for a given language and category pair, we grouped sentences of similar size together by sorting them by length and placing in 20 buckets of equal size, (iv) we sampled 50 sentences from each bucket ensuring that each sentence has exactly one entity of the expected category (using regular expressions listed in Appendix~\ref{appendix:regex}) and has no grammar errors according to \texttt{language tool}~\cite{10.5555/1815744.1815745}~\footnote{\url{https://pypi.org/project/language-tool-python/}}. The buckets provide a diverse length representation, (v) Finally, we combined the selected samples forming a set of 1,000 examples (50 examples $\times$ 20 buckets) for each language and category pair.

This procedure applied to each language and category created a high-quality dataset of 36,000 examples (1,000 examples $\times$  9 categories $\times$ 4 languages). The dataset is published online~\footnote{\url{https://github.com/laniqo-public/do-not-change-me}}.

\section{Explorative data analysis}

\begin{figure}[h]
    \centering
    \includegraphics[width=0.48\textwidth]{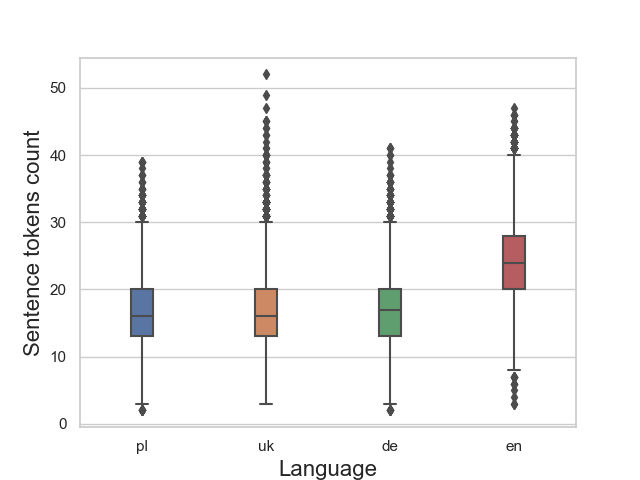}
    \caption{Number of NLTK~\cite{bird2009natural} tokens in sentences expressed in a given language.}
    \label{fig:sentence_lang_tokens_distribution}
\end{figure}

The average sentence in our dataset consists of 18.71 ($\pm$ 6.81 std. dev.) tokens\footnote{Tokens were generated using the NLTK's~\cite{bird2009natural} \texttt{word\_tokenize} function}. As can be seen in Figure~\ref{fig:sentence_lang_tokens_distribution}, Polish, Ukrainian, and German sentences tend to be similar in length -- with an average of almost 17 tokens. Meanwhile English sentences are longer, with the average of 23.99 tokens. This discrepancy arises from Gemma's tendency to include English translations for short inputs and non-English target languages due to the \texttt{min\_tokens=16} constraint. In these cases, our filtering procedure described in Section~\ref{sample_selection} is applied, making the average sentence shorter.

There is a greater variability in the number of tokens in sentences when analyzing texts by category. These values, presented in Appendix~\ref{appendix:figures}, Figure~\ref{fig:category_tokens_sentences}, show that sentences introducing phone numbers are on average the longest with 23.21 $\pm$ 6.68 tokens. The shortest sentences are related to emojis, with 13.06 $\pm$ 3.94 tokens. 

Considering the lengths of the entities in characters, we see that the average entity consists of 17.91 $\pm$ 9.99 characters. Contrary to general sentence lengths, per-language entity length averages are very similar to each other, with English entities having 18.13 $\pm$ 10.48 characters, German -- 18.01 $\pm$ 9.64, Polish -- 17.93 $\pm$ 9.81, and Ukrainian -- 17.58 $\pm$ 10.01. The shortest entities are emojis, usually consisting of only 1 character, while the longest are URLs, emails, IBANs, and phone numbers. The longest entity is an English URL address consisting of more than 100 characters.

\begin{figure*}[h]
    \centering
    \includegraphics[width=0.9\textwidth]{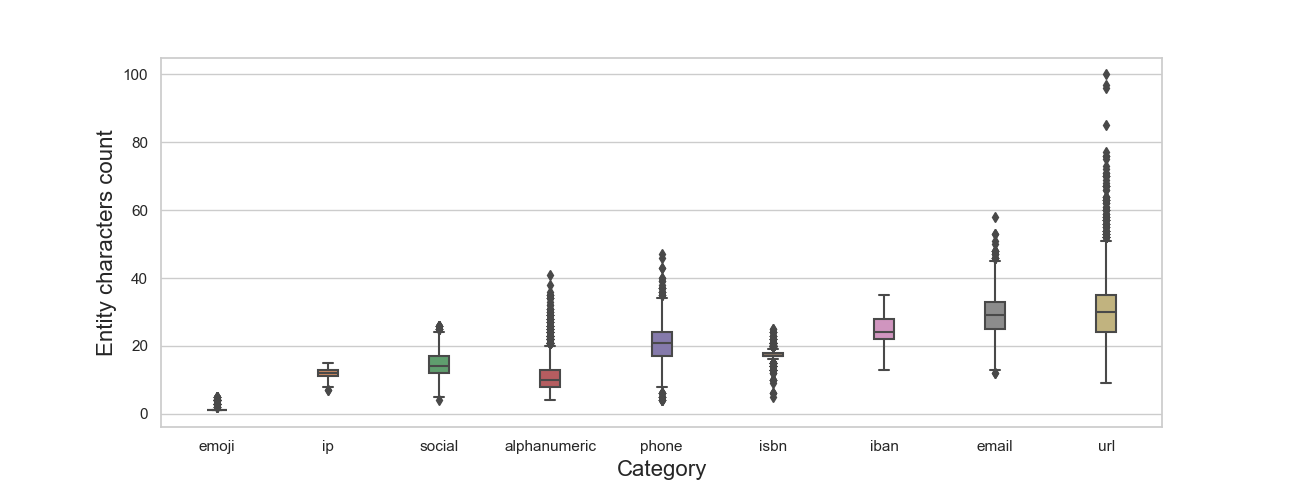}
    \caption{Distribution of the number of characters for each entity category.}
    \label{fig:entity_characters_distribution}
\end{figure*}

\section{Methodology}
To understand the quality of the entity transfer, we applied the following methodology:
(i) we chose a set of NMT models to translate between supported languages, (ii) we used each selected model to translate each sentence from our dataset to all languages considered, and (iii) we extracted entities from source and target sentences and compared them in terms of the Levenshtein distance. The following subsections describe these steps in detail.


\subsection{Models selection}
We selected the most popular NMT models from the Huggingface repository, searching for models that support all the languages considered (English, German, Polish, and Ukrainian), and those that fit consumer GPUs (not bigger than 9B parameters). The list of the selected models is given in Table~\ref{tab:models}. 

Since OPUS-MT models are unidirectional, with specialized models translating between a given language pair, we selected 12 OPUS models that support all possible language pairs considered.

Moreover, we selected Google Translate as a reference as it is one of the leading commercial translation services. 

\begin{table}[]
\small
\centering

\caption{Models selected for experiments.}
\begin{tabular}{|c|c|}
\hline 
  Model           & \# of Params  \\ \hline 
  OPUS~\footnote{\url{https://hf.co/Helsinki-NLP/opus-mt-en-de}} & 12$\times$75M \\ 
  mBART~\footnote{\url{https://hf.co/facebook/mbart-large-50-many-to-many-mmt}} & 611M \\
  NLLB~\footnote{\url{https://hf.co/facebook/nllb-200-3.3B}} & 3,300M \\
  M2M100~\footnote{\url{https://hf.co/facebook/m2m100_1.2B}} & 1,200M \\
  EuroLLM 9B~\footnote{\url{https://hf.co/utter-project/EuroLLM-9B-Instruct}} & 9,000M \\
  MADLAD 7B~\footnote{\url{https://hf.co/google/madlad400-7b-mt}} & 7,000M \\
  SeamlessM4T~\footnote{\url{https://hf.co/facebook/seamless-m4t-v2-large}} & 2,300M \\ \hline
  Google Transl. & no data \\ \hline
\end{tabular}
\label{tab:models}
\end{table}


\subsection{Translation procedure}

For each model, language, and category, we translated sentences from a given language to all other supported languages. We used the Huggingface Transformers\footnote{https://huggingface.co/} and vLLM\footnote{https://github.com/vllm-project/vllm} libraries for inference. In this way, we generated 108,000 translations (3$\times$36,000). 

For Google Translate, we used the Google Docs translation feature to translate sentences in batches. For each language and category, we concatenated all sentences related to a given language and category into one document using double newlines to separate sentences and translated these documents. In this way, we reduced the number of requests. 

The generated translations  are published online\footnote{\url{https://github.com/laniqo-public/do-not-change-me}}.

\subsection{Source and target sentence entity comparison}
For each source and translation (target), we used the hand-crafted regular expressions listed in Appendix~\ref{appendix:regex} to detect if the desired entity is transferred correctly. Due to the selection process, each source sentence contains exactly one entity of a given type. Thanks to regular expressions, we checked if entities of the same type are detected in both source and target, and measure differences between them if needed, using Levenshtein distance. To add an additional perspective on the general quality of the translations, we used the CometKiwi\footnote{\scriptsize Python3.12.7|Comet2.2.4|fp32|Unbabel/wmt22-cometkiwi-da|r0}~\cite{rei-etal-2022-cometkiwi} model.



\section{Results}
\label{sec:results}

We analyzed the results from various perspectives, as described in the following paragraphs.

\paragraph{Per-category accuracy} For each category considered, we examined each model's entity transfer accuracy averaging scores over all language directions with equal weights. In this experiment, a true positive is generated when a regular expression related to the expected category detects exactly the same sequence of characters on the source and target sides.

Table~\ref{tab:accuracy_category} summarizes this scenario and shows that the best-performing models are \textbf{EuroLLM 9B}, \textbf{Google Translate}, and \textbf{MADLAD}. 
The worst-performing models are \textbf{OPUS}, \textbf{SeamlessM4T}, and \textbf{NLLB}. Emoji was the category where most models struggled. Five models: \textbf{OPUS}, \textbf{MBART}, \textbf{SeamlessM4T}, \textbf{NLLB} and \textbf{M2M100} achieved an accuracy of less than 5.5\% in this category. The three easiest-to-transfer categories for all models are: IP (average over models: 91.79), Phone (91.29), and ISBN (90.69). These scores suggest that all those models are able to transfer sequences of numbers relatively well.

\begin{table*}[h]
\small
\centering
\caption{Accuracy averaged over all language pairs for each category. Values in bold are the highest, while underlined are the worst.}
\label{tab:accuracy_category}
\begin{tabular}{|r|>{\columncolor[gray]{0.8}}c|c|c|c|c|c|c|c|}
\hline 
Category     & Google T.. & OPUS  & MADLAD & MBART & SeamlessM4T & NLLB  & EuroLLM 9B & M2M100\\ \hline
alphanum & 87.39            & \underline{71.67} & 81.68  & 86.65 & 78.82       & 83.31 & 87.77 & \textbf{88.54} \\
email        & 85.66           & \underline{30.31} & 85.52  & 89.15 & 82.78       & 84.72 & \textbf{92.12} & 90.14 \\
emoji        & \textbf{98.59}            & \underline{0.02}  & 81.64  & 2.1   & 0.67        & 5.3   &  96.78 & 4.18\\
iban         & 95.45            & \underline{40.45} & 79.53  & 84.87 & 80.47       & 88.96 &  \textbf{98.55} & 94.86 \\
ip           & 98.61            & \underline{58.12} & 94.49  & 97.82 & 91.28       & 95.62 & \textbf{99.58} & 98.78 \\
isbn         & \textbf{99.16}            & \underline{69.91} & 89.22  & 98.51 & 76.54       & 97.05 & 98.41  & 96.7 \\
phone        & 98.38            & \underline{68.08} & 90.52  & 96.45 & 83.22       & 96.24 &  98.36  & \textbf{99.07} \\
social       & 88.08            & \underline{34.51} & 84.37  & 88.79 & 77.50       & 75.04 & \textbf{96.23} & 72.00\\
url          & 88.52            & \underline{38.09} & 81.96  & 90.43 & 50.92       & 79.11 & \textbf{95.20}  &  86.09 \\ \hline
macro avg. & 93.32 &	\underline{45.68} &	85.44 &	81.64 &	69.13 & 78.37 & \textbf{95.89}  &	81.15\\ \hline 
\end{tabular}
\end{table*}
\begin{table*}[h]
\small
\centering
\caption{Accuracy macro-averaged over categories for each direction. Values in bold are the highest, while underlined are the worst.}
\label{tab:accuracy_direction}
\begin{tabular}{|r|>{\columncolor[gray]{0.8}}c|c|c|c|c|c|c|c|}
\hline 
Direction & Google T. & OPUS  & MADLAD & MBART & SeamlessM4T & NLLB  & EuroLLM 9B & M2M100 \\ \hline 
de $\rightarrow$ en   & 90.31            & 68.74 & 92.84 & 82.91 & \underline{66.06}       & 74.56 &  \textbf{95.75}  & 78.98  \\
de $\rightarrow$ pl   & 89.52            & \underline{58.33} & 88.92  & 82.25 & 84.20       & 75.42 &    \textbf{94.92}   & 79.27 \\
de $\rightarrow$ uk   & 90.09            & \underline{14.60} & 85.61  & 80.52 & 83.04       & 76.25 &    \textbf{94.62}  &  76.27\\
en $\rightarrow$ de   & 96.92            & 81.27 & 95.36  & 87.14 & \underline{74.01}       & 85.41 &    \textbf{97.18} &  86.66 \\
en $\rightarrow$ pl   & 97.04            & 77.58 & \textbf{97.88}  & 88.07 & \underline{74.23}       & 86.31 &    97.75 & 87.70 \\
en $\rightarrow$ uk   & \textbf{98.85}            & \underline{21.42} & 96.07  & 86.85 & 38.45       & 85.26 &    97.08  &  86.60 \\
pl $\rightarrow$ de   & 90.29            & \underline{54.56} &  88.37 & 77.54 & 83.17       & 69.98 &    \textbf{94.70}  & 74.35  \\
pl $\rightarrow$ en   & 91.70            & 59.44 &  94.69 & 81.35 & \underline{31.89}       & 68.85 &  \textbf{95.49}     & 74.67 \\
pl $\rightarrow$ uk   & 90.86            & \underline{25.65} &  47.01 & 79.33 & 82.57       & 70.95 &    \textbf{95.17}   & 74.60 \\
uk $\rightarrow$ de   & 93.22            & \underline{21.61} &  86.23 & 76.23 & 84.66       & 82.59 &    \textbf{95.21}   & 84.34 \\
uk $\rightarrow$ en   &\textbf{96.60}            & \underline{33.85} &  92.24 & 80.78 & 42.43       & 82.32 &     96.48   & 84.96 \\
uk $\rightarrow$ pl   & 94.41            & \underline{31.15} &  60.02 & 76.69 & 84.91       & 82.58 &    \textbf{96.29}   & 85.45 \\ \hline 
macro avg. &	93.32 & \underline{45.68}	& 85.44	& 81.64	& 69.13	& 78.37	& \textbf{95.89} & 81.15 \\ \hline 
\end{tabular}
\end{table*}

\paragraph{Per-direction accuracy} Similarly to per-category analysis, we also grouped the results by translation direction, macro-averaging the scores over all categories considered. The reason for this analysis is to understand whether all the language pairs considered are handled with similar quality. 

The corresponding statistics are presented in Table~\ref{tab:accuracy_direction}. The highest quality overall is observed for the en $\rightarrow$ uk direction, with an average accuracy of 98.85. The lowest accuracy (14.60) is observed for de $\rightarrow$ uk for the \textbf{OPUS} model. The average accuracy over all models and directions is equal to 78.65 ($\pm$ 19.3 of std. dev.). Averaging over all models, the best quality directions are: en $\rightarrow$ pl (88.32 $\pm$ 9.06), en $\rightarrow$ de (87.99 $\pm$ 8.18), de $\rightarrow$ pl (81.6 $\pm$ 11.25) and de $\rightarrow$ pl (81.27 $\pm$ 11.13). On average, the worst quality directions are pl $\rightarrow$ uk (70.67 $\pm$ 23.38), pl $\rightarrow$ en (74.76 $\pm$ 21.6), and de $\rightarrow$ uk (75.13 $\pm$ 25.27). However, it should be noted that the weak average scores involving Ukrainian language are due to \textbf{OPUS} and \textbf{SeamlessM4T} models, which handle this language relatively poorly in the context of entity transfer. While, according to Table~\ref{tab:accuracy_direction}, among all directions, the average accuracy for en $\rightarrow$ uk direction is the highest for \textbf{Google Translate}, second highest for \textbf{MADLAD}, and third highest on \textbf{MBART}, \textbf{NLLB}, \textbf{EuroLLM}, and \textbf{M2M100}, it is ranked 11th out of 12 possible places for \textbf{OPUS} and \textbf{SeamlessM4T}. 

\begin{table*}[]
\tiny
\centering
\caption{Prompts used for EuroLLM.}
\label{tab:euro_prompt}
\begin{tabular}{|r|p{14cm}|}
\hline 
Prompt id     & Text \\ \hline 
GENERIC & <|im\_start|>system
You are a professional \{source\_language\}
to \{target\_language\} translator.Your goal
is to accurately convey the meaning and 
nuances of the original \{source\_language\} 
text while adhering to \{target\_language\} 
grammar, vocabulary, and cultural sensitivities. 
Return only translation without any prefixes and explanations.
<|im\_end|>
<|im\_start|>user 
Translate the following \{source\_language\} source text to \{target\_language\}:
\{source\_language\}: \{source\}
\{target\_language\}: <|im\_end|>
<|im\_start|>assistant \\ \hline 
FOCUSED & <|im\_start|>system
You are a professional \{source\_language\} to \{target\_language\} translator. Your goal is to accurately convey the meaning and nuances of the original \{source\_language\} text while adhering to \{target\_language\} grammar, vocabulary, and cultural sensitivities.
\textbf{Translate the provided text while ensuring that all non-translatable elements, such as numbers, email addresses, alphanumeric strings, emoticons, and similar elements, remain unchanged in the translated text.}
Return only translation without any prefixes and explanations.
<|im\_end|>
<|im\_start|>user
Translate the following \{source\_language\} source text to \{target\_language\}:
\{source\_language\}: \{source\}
\{target\_language\}: <|im\_end|>
<|im\_start|>assistant \\ \hline 

\end{tabular}
\end{table*}

\begin{table*}[h]

\centering
\small
\caption{Average change when using large models and smaller ones. For EuroLLM, 2 prompts are evaluated.}
\label{tab:size_model}
\begin{tabular}{|r|>{\columncolor[gray]{0.8}}c|c|c|c|c||c|c|}
\hline 
             &    Reference              & \multicolumn{2}{|c|}{Focused prompt}   & \multicolumn{2}{c|}{Generic prompt} & \multicolumn{2}{c|}{}          \\ \hline 
Category     & Google T.. & Euro 1.7B  & Euro 9B  & Euro 1.7B  & Euro 9B & MADLAD 3.3B & MADLAD 7B \\ \hline
alphanum     & 87.49  & 83.03 & \textbf{87.77}  & \underline{82.56} & 86.42 & 81.25 & 81.68  \\
email        & 85.66 &  71.56 & \textbf{92.12} & \underline{71.06} & 90.61  & 86.47 & 85.52\\
emoji        & \textbf{98.59} &  89.43 & 96.78  & \underline{84.00}  & 93.19 & 75.95 & 81.64 \\
iban         & 95.45 &  92.64 & \textbf{98.55}  & \underline{91.80} & 98.12 & 80.53 & 79.53 \\
ip           &  \underline{98.61}  &  98.86 & \textbf{99.58} & 98.77 & 99.35 & 96.72 & 94.49\\
isbn         & \textbf{99.16}  & 95.73  & 98.41 & \underline{95.06}  & 98.19 & 88.17 & 89.22\\
phone        & \textbf{98.38}  &    97.35  & 98.36  & \underline{97.25} & 98.08  & 88.48 & 90.52 \\
social       & 88.08  &  \underline{84.74}  & \textbf{96.23} & 85.27 & 95.29 & 81.41 & 84.37\\
url          & 88.52  & 88.19  & \textbf{95.20}  & \underline{87.93} & 94.26 & 85.23 & 81.96 \\ \hline
macro avg. & 93.32 & 89.06 &	\textbf{95.89} & \underline{88.19} & 94.83	 & 84.91 &  85.44\\ \hline 
\end{tabular}
\end{table*}

\paragraph{Error distribution} 

The accuracy scores analyzed in the previous paragraphs give only partial information on what happens to entities during translations, as the accuracy only checks whether exactly the same entity is detected on the source and target sides. To address this problem, we analyzed the distribution of errors in terms of the Levenshtein distance. For each model, we checked what percentage of errors are due to the lack of an entity of a given type detected on the target side (no-match case), or matching an entity of a given type but differing in the sequences detected. In that case, we measured the edit distance between the expected and the observed entities (between the source and target entities). The summary of this experiment is presented in Appendix~\ref{appendix:figures}, Figure~\ref{fig:errors}. For 6 models (\textbf{OPUS}, \textbf{MADLAD}, \textbf{MBART}, \textbf{SeamlessM4T}, \textbf{NLLB}, and \textbf{M2M100}), the no-match category is by far the most dominant. This is mainly due to the emoji category, which according to Table~\ref{tab:accuracy_category} has very low scores assigned to exactly these models (except \textbf{MADLAD}). The low scores for emojis are due to the lack of model's support for byte-level tokenization.

A deeper analysis reveals that these six models indeed produce most of the no-match errors from emojis: 85.84\% of \textbf{MBART} no-match errors are due to emojis, 85.58\% for \textbf{M2M100}, 68.99\% for \textbf{NLLB}, 60.78\% for \textbf{SeamlessM4T}, 41.16\% for \textbf{OPUS}, and 25.37\% for \textbf{MADLAD}. Models that are not dominated by no-match errors (\textbf{Google Translate} and \textbf{EuroLLM}, have a relatively low percentage of emojis in the no-match category: 12.5\% and 26.73\%, respectively. For comparison, Figure~\ref{fig:errors_dist_no_emoji} in Appendix~\ref{appendix:figures} represents the error distribution without the Emoji category. In every scenario, small differences (e.g., 1 or 2 characters) are more likely than larger ones (e.g., 3 or 4). The highest chance of one-character errors is observed for \textbf{Google Translate}. Models, such as \textbf{EuroLLM}, \textbf{MADLAD}, or \textbf{Google Translate} have a relatively high number of large differences (Levenshtein distance > 5).

Table~\ref{tab:lev_analysis} provides information on the category that each model has the most problems with, listing the categories with the highest number of errors per each model considered. This analysis is performed separately for edit distances equal to one, two, and larger than 5. As can be seen, the most frequent differences by one character can be observed for IBANs and social handlers (2/8 models struggled with each of those categories). Larger differences, with edit distance = 2, are more common among IBANs and ISBNs (three models struggled with IBANs and other three with ISBNs). When considering edit distances larger than 5, social handlers are the most problematic for 3 models, and alphanumeric sequences for 2 models. Some recurring problems that are observed among translations are: repeating the same phrase over and over, dropping a subset of characters, translating fragments of entities, and omitting an entity. Table~\ref{tab:google_translate_errors} presents a subset of problems observed for \textbf{Google Translate}.

\begin{table*}[h]
\caption{Categories generating highest number of errors in relation to the edit distance considered.}
\label{tab:lev_analysis}
\centering
\small
\begin{tabular}{|l|l|l|l|l|l|l|}
\hline 
                 & \multicolumn{2}{c|}{Lev. = 1} & \multicolumn{2}{c|}{Lev. = 2} & \multicolumn{2}{c|}{Lev. \textgreater 5} \\
Model            & category       & \# errors   & category     & \# errors     & category             & \# errors        \\ \hline
Google Translate & IBAN         & 339         & e-mail       & 195           & alphanumeric         & 736             \\
OPUS             & phone         & 1074        & ISBN         & 1479          & social               & 2275             \\
MADLAD           & IBAN           & 381         & IBAN         & 166           & phone         & 955             \\
MBART            & alphanumeric   & 288         & IBAN         & 178           & IBAN         & 829             \\
SeamlessM4T      & ISBN           & 677         & ISBN         & 774           & URL                  & 3239             \\
NLLB             & social         & 303         & IBAN         & 290           & social         & 1574             \\
EuroLLM          & e-mail         & 94         & ISBN         & 120           & alphanumeric         & 786             \\
M2M100           & social         & 314         & URL          & 239           & social         & 1364            \\ \hline
\end{tabular}
\end{table*}




\paragraph{Prompt selection for EuroLLM}
We experimented with two types of prompts for inference using \textbf{EuroLLM} models: generic and focused ones, as described in Table~\ref{tab:euro_prompt}. The focused prompt differs from the generic one by adding a sentence requesting that the entities representing categories considered should remain unchanged in translated sentences. The results, presented in Table~\ref{tab:size_model}, indicate that the focused prompt significantly improves the handling of entities for both versions of \textbf{EuroLLM} (1.7B and 9B). Additionally, the results of CometKiwi, shown in Table~\ref{tab:kiwi}, demonstrate that inference with the focused prompt also enhances translation quality. Consequently, we reported the results of \textbf{EuroLLM} using the focused prompt in all experiments.

\paragraph{Model size vs. accuracy}

The models analyzed differ both in size and the number of supported language directions. As bigger models are frequently linked to higher quality in various NLP tasks, it may be interesting to evaluate the relationship between the model size and the number of parameters in comparison to the average accuracy. With the details on model sizes and parameters provided in Table~\ref{tab:params_analysis}, we can observe that the biggest models, namely \textbf{EuroLLM (9B)} and \textbf{MADLAD (7B)} are indeed of the best quality and the smallest ones, the \textbf{OPUS} family, is the worst. Interestingly, those large models are of the highest quality despite the large number of language directions supported. \textbf{OPUS}, with the highest number of parameters per a direction (75M) is the worst-quality model, which may indicate that multilingual models do not lose their entity transfer abilities with more languages. We also directly compared two of the largest and best performing models -- \textbf{EuroLLM (9B)} and \textbf{MADLAD (7B)} with their smaller counterparts -- \textbf{EuroLLM (1.7B)} and \textbf{MADLAD (3B)}. As presented in Table~\ref{tab:size_model}, the larger versions perform better. \textbf{EuroLLM} 9B beats the 1.7B version in the case of every category, with the average accuracy more than 6 percentage points higher in the case of the 9B model. Larger \textbf{MADLAD}, on the other hand, beats the smaller version in 5 out of 9 categories, with similar scores in 3 other categories, and a drop of more than 3 percentage points compared to the 3B model on URL addresses. 

\paragraph{Relationship between accuracy and average entity length} Our intuition tells us that the probability of errors may depend on the number of (sub)tokens a given entity is tokenized into by a given model. As different models use different tokenizers, they can split entities in a different manner. In Table~\ref{tab:params_analysis}, we put together the accuracies of models with the average entity lengths measured in tokens generated with a tokenizer of a given model. The Spearmans's rank correlation measured between these values tells us that there is a non-significant medium positive relationship between the accuracy and average entity (sub)tokens number ($r_s$ = 0.3929, p = 0.395). The same conclusion is reached for Pearson's correlation with $r$ = 0.4317 and p = 0.334. This observation may indicate that models splitting entities into a longer sequence of smaller (sub)tokens deal better with the transfer task, whereas models that chunk entities into bigger portions struggle in that context. For example, \textbf{EuroLLM} tokenizes an example ISBN number \texttt{0176 7890 1234 5678 9012 3456} into 30 tokens:  "\_",  "0",  "1",  "7",  "6",  "\_",  "7",  "8",  "9",  "0",  "\_",  "1",  "2",  "3",  "4",  "\_",  "5",  "6",  "7",  "8",  "\_",  "9",  "0",  "1",  "2",  "\_",  "3",  "4",  "5",  "6", while \textbf{OPUS} generates only 12 tokens: "\_01", "76", "\_78", "90", "\_12", "34", "\_56", "78", "\_90", "12", "\_34", "56". As can be seen, \textbf{OPUS}, instead of relating sequences of digits to ISBNs (as \textbf{EuroLLM} did), needs to consider possible pairs of digits, which are more numerous. 

Based on the token length distribution presented in Figure~\ref{fig:entity_characters_distribution}, we selected categories with high average length and analyzed the correlation between the number of entity's (sub)tokens and the likelihood of observing an error in an entity. The results presented in Table~\ref{tab:pearson_categories} show that there is a positive correlation between URL, alphanumeric, and e-mail lengths and error likelihood, ranging from very small values (\textbf{EuroLLM} and URL category -- Pearson's coeff. = 0.179 with p-value = 0.303) and very high ones (\textbf{OPUS} with e-mail and \textbf{URL} categories -- 0.822 with p-value = 4.9e-06 and 0.724 with p-value = 5.94e-08, respectively). This analysis shows that some models, e.g.,  \textbf{OPUS} and \textbf{MBART} are more prone to making errors in longer sequences, whereas, e.g., \textbf{EuroLLM} model is much more robust.

\begin{figure*}[h]
    \centering
    \includegraphics[width=0.85\textwidth]{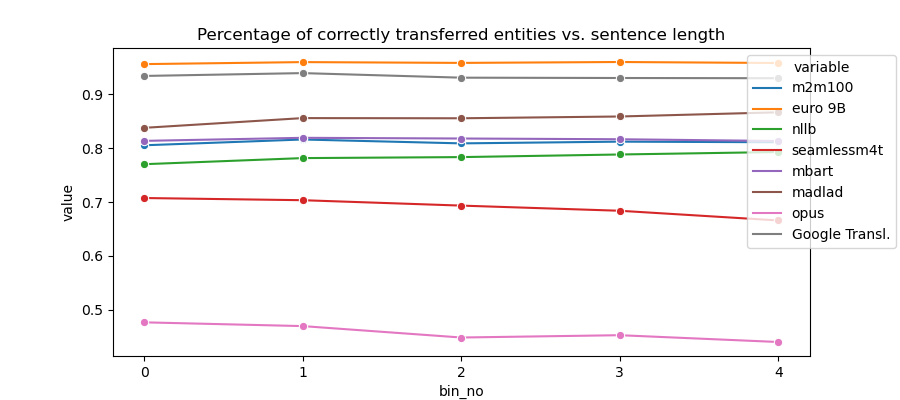}
    \caption{Percentage of correctly transferred entities among different sentence lengths.}
    \label{fig:bins_ok}
\end{figure*}

\begin{table*}[h]
\centering
\small
\caption{Pearson's correlation coefficient between the number of subtokens in an entity and the likelihood of observing an error in an entity}
\label{tab:pearson_categories}
\begin{tabular}{|c||c|c||c|c||c|c|c|c|}
\hline 
 Model & \multicolumn{2}{c||}{Alphanumeric} & \multicolumn{2}{c||}{email} & \multicolumn{2}{c|}{URL} & \multicolumn{2}{c|}{Phone} \\ 
    & Pearson's c.  & p-value & Pearson's c.  & p-value & Pearson's c.  & p-value & Pearson's c.  & p-value \\ \hline 
  EuroLLM  & 0.344 & 0.1 & 0.401 & 0.111 & 0.179 & 0.303 & -0.07 & 0.665 \\ 
  MADLAD  & 0.578 & 0.003 & 0.453 & 0.059 & 0.179 & 0.29 & -0.187 & 0.241  \\ 
  OPUS  & 0.61 & 0.002 & 0.822 & 4.9e-06 & 0.724 & 5.94e-08 & 0.624 & 0.001\\
  M2M100  & 0.479 & 0.024 & 0.572 & 0.01 & 0.046 & 0.785 & -0.277 & 0.224\\
  SeamlessM4T  & 0.402 & 0.071 & 0.696 & 0.001 & 0.397 & 0.018 & 0.289 & 0.192\\
  NLLB  & 0.473 & 0.026 & 0.47 & 0.049 & 0.194 & 0.263 & 0.192 & 0.418\\
  MBART  & 0.814 & 1.27e-05 & 0.603 & 0.01 & 0.39 & 0.023 & -0.351 & 0.13\\ \hline 
\end{tabular}
\end{table*}

\paragraph{Context length vs. accuracy}
Each entity is mentioned within a sentence, which can be considered its context. To understand whether longer sentences are more beneficial, we checked if the number of (sub)tokens in the sentence influences the accuracy of transfer. The intuition is that longer sentences should make entities less ambiguous, leading the model to realize that those entities should be transferred without changes. To measure the degree of this relationship, the following procedure was applied: we sorted source sentences according to lengths and split them into five bins of equal size so that texts of similar lengths are in the same bin. 
Then, for each bin, we measured the average accuracy of a given model, and the ratios of situations where entities are modified (Levenshtein > 0) or are not transferred at all (no-match errors). Figure~\ref{fig:bins_ok} shows the percentage of correctly transferred entities for each bin. Figures~\ref{fig:bins_nomatch} and~\ref{fig:bins_modified} in Appendix~\ref{appendix:figures} show distributions of non-matched and modified entities in target sentences, respectively. As can be seen, there is no visible relationship between the length of the sentence and the accuracy of the entity transfer. For some models, e.g., \textbf{OPUS}, the quality even decreases with increasing sentence length.



\begin{table*}[h!]

\centering
\small
\caption{Relationship between the accuracy of a model and selected model characteristics.}
\label{tab:params_analysis}
\begin{tabular}{|c|>{\centering}p{1.5cm}|c|>{\centering}p{2cm}|>{\centering}p{2cm}|c|}
\hline 
  Model           &  Num. of params (M)  & Supported languages  & params per language (M) & avg. tokenized entity length & Average accuracy        \\ \hline 
  EuroLLM 9B & 9,000 & 1,225 ($35^2$) & 7.35 & 13.55 &  95.89 \\ 
  MADLAD & 7,000 & 175,561 ($419^2$) & 0.04 & 13.42 & 85.44 \\  
  OPUS & 75 & 1 ($1^1$) & 75  & 10.31 & 45.68\\ 
  M2M100 & 1,200 & 10,000 ($100^2$) & 0.12 & 9.64 & 81.15 \\  
  SeamlessM4T & 2,300 & 9,216 ($96^2$) & 0.25 & 9.55 & 69.13 \\ 
  NLLB & 3,300 & 40,000 ($200^2$) & 0.083 & 9.51 & 78.37 \\ 
  MBART & 611 & 2500 ($50^2$) & 0.244 & 8.59 &  81.64 \\ \hline
\end{tabular}
\end{table*}

\paragraph{General translation quality}
The CometKiwi translation evaluation is presented in Table~\ref{tab:kiwi} and described in Appendix~\ref{app:kiwi}. The scores range from 0.7 to 0.73 in most scenarios, with \textbf{Google Translate} -- the top-rated model -- assigned a score of 0.76. This means that, in general, the overall translation quality was good for all models considered. 

\section{Conclusions}
In this paper, we find that modern medium-sized LLMs such as \textbf{EuroLLM}, despite being only partially trained on parallel corpora, may excel in terms of entity transfer quality and can reach the quality similar to \textbf{Google Translate}. As the best performing model is \textbf{EuroLLM 9B}, it is the answer to \textbf{RQ3}. In contrast, the smallest  \textbf{OPUS}-related models are scored as the worst. This may be due to two factors: on the one hand, bigger models may learn more about the world, but also they may learn better token representations using fine-grained tokenization as discussed in Section~\ref{sec:results}. We showed that increasing the length of the sentence does not correlate with the transfer accuracy, thus, the context information does not necessarily help to guide a model to transfer a given sequence without modifications. Considering \textbf{RQ2}, the manual analysis of entities transferred by models with high Levenshtein distances reveals that entities are frequently partially translated, differing in some numbers, or repeated in fragments. Also, for some models (e.g., \textbf{OPUS}, \textbf{MBART}), the longer entities in terms of subtokens increase the probability of generating an error. This is in line with the intuition that it may be harder to transfer long sequences without errors as compared to the short ones. While most of the categories considered are decently transferred by most of the models considered, emoji is a big struggle -- \textbf{OPUS}, \textbf{MBART}, \textbf{SeamlessM4T}, \textbf{NLLB}, and \textbf{M2M100} cannot transfer this category correctly (which answers \textbf{RQ1}). 




\bibliography{mtsummit25}

\appendix

\paragraph{Sustainability statement}
The experiments were performed on a single PC with one GeForce RTX 4090 GPU. The total time for the experiments was equal to approximately 12 hours of translation time, which is related to 1.56 kg of CO2eq according to \texttt{\url{https://mlco2.github.io/impact/}}.

\section{Regular expressions for category detection}
\label{appendix:regex}
List of regular expressions utilized for identifying non-translatable units:

\textbf{Alphanumeric}: \begin{verbatim}
  \b[\p{N}\p{L}][\p{N}\p{L}\p{P}]*
  (\p{L}\p{N}|\p{N}\p{L})
  [\p{N}\p{L}\p{P}]*[\p{N}\p{L}]\b
\end{verbatim}

\textbf{E-mail}: \begin{verbatim}
  \b[\p{L}\p{N}._%+-]+@[\p{L}\p{N}.-]+
  \.[\p{L}]{2,}\b
    \end{verbatim}

\textbf{IBAN}: \begin{verbatim}
    \b([A-Z]{2})[ \-]?([0-9]{2})[ \-]?
    ([A-Z0-9]{9,30})\b
\end{verbatim}

\textbf{IP}: \begin{verbatim}
    \b\d{1,3}\.\d{1,3}\.
    \d{1,3}\.\d{1,3}\b
\end{verbatim}

\textbf{ISBN}: \begin{verbatim}
    \b(?:ISBN(?:-13)?:?\ )?(?=[0-9]{13}$|
    (?=(?:[0-9]+[-\ ]){4})[-\ 0-9]{17}$)
    97[89][-\ ]?[0-9]{1,5}[-\ ]?[0-9]+
    [-\ ]?[0-9]+[-\ ]?[0-9]\b
\end{verbatim}

\textbf{Phone}: \begin{verbatim}
    \b[\d\+\/\=\%\^\(\)\[\]\{\}]
    [\d\., \+\:\-*\/\=\%\^\(\)\[\]\{\}]
    {2,}
    [\d\+\:\-*\/\=\%\^\(\)\[\]\{\}]\b
\end{verbatim}

\textbf{Social handler}: \begin{verbatim}
    \@[0-9_.\p{L}]{2,24}[0-9_\p{L}]\b
\end{verbatim}

\textbf{URL}: \begin{verbatim}
    \b((imap|s3|file|ftp|https?):\/\/
    [\p{L}\p{N}_-]+
    (\.[-_/?=\p{L}\p{N}]+){1,15}|
    \d{1,3}\.\d{1,3}\.\d{1,3}\.\d{1,3}|
    www\.[\p{L}\p{N}_-]+
    (\.[-_/?=\p{L}\p{N}]+){1,15})\b
\end{verbatim}

\section{Prompts used to generate samples}
\label{appendix:generation}
\begin{itemize}
    \item \textbf{Alphanum}: Write me a random and creative sentence in [LANGUAGE] that includes a sequence consisting of multiple digits and letters longer than 5 characters.
    \item \textbf{E-mail}: Write me a random and creative sentence in [LANGUAGE] that includes a random email address.
    \item \textbf{Emoji}: Write me a random and creative sentence in [LANGUAGE] that includes a random emoji.
    \item \textbf{IBAN}: Write me a random and creative sentence in [LANGUAGE] with a sequence including an artificial IBAN number in IBAN format.
    \item \textbf{ISBN}: Write me a random and creative sentence in [LANGUAGE] with a sequence including an artificial ISBN number in ISBN format.
    \item \textbf{IP}: Write me a random and creative sentence in [LANGUAGE] that includes a random IP number.
    \item \textbf{Phone}: Write me a random and creative sentence in [LANGUAGE] that includes a long random phone number.
    \item \textbf{Social}: Write me a random and creative sentence in [LANGUAGE] that includes a social media handler starting with the @ sign (e.g., Twitter, Instagram).
    \item \textbf{URL}: Write me a random and creative sentence in [LANGUAGE] that includes a random URL address.
\end{itemize}

\section{Detailed CometKiwi evaluation}
\begin{table*}[h!]
\caption{CometKiwi evaluation of translation accuracy across various models, language pairs, and data categories, showcasing macro-averaged performance by language and prompt-specific configurations.}
\resizebox{\textwidth}{!}{%
    \centering
\begin{tabular}{|l|c|c|c|c|c|c|c|c|c|c|c|c|c|}
\hline
                                                              & Google T. & OPUS   & \begin{tabular}[c]{@{}c@{}}MADLAD \\ 3.3B\end{tabular} & \begin{tabular}[c]{@{}c@{}}MADLAD \\ 7B\end{tabular} & MBART  & SeamlessM4T & NLLB   & M2M100 & \begin{tabular}[c]{@{}c@{}}EuroLLM \\ 1.7B\\ generic \\ prompt\end{tabular} & \begin{tabular}[c]{@{}c@{}}EuroLLM \\ 1.7B\\ focused \\ prompt\end{tabular} & \begin{tabular}[c]{@{}c@{}}EuroLLM\\  9B \\ generic\\  prompt\end{tabular} & \begin{tabular}[c]{@{}c@{}}EuroLLM \\ 9B\\ focused\\  prompt\end{tabular} & \begin{tabular}[c]{@{}c@{}}macro avg. \\ by lang and\\ category\end{tabular} \\ \hline
de-en                                                         & 0.795     & 0.7861 & 0.7852                                                 & 0.7863                                               & 0.7839 & 0.7899      & 0.787  & 0.7775 & 0.7787                                                                      & 0.7796                                                                      & 0.781                                                                      & 0.7822                                                                    & 0.7844                                                                       \\
de-pl                                                         & 0.7474    & 0.6853 & 0.7042                                                 & 0.71                                                 & 0.6966 & 0.6859      & 0.6968 & 0.7047 & 0.7033                                                                      & 0.7022                                                                      & 0.7056                                                                     & 0.7085                                                                    & 0.7042                                                                       \\
de-uk                                                         & 0.7529    & 0.6511 & 0.6816                                                 & 0.6918                                               & 0.6676 & 0.6745      & 0.6742 & 0.6882 & 0.6895                                                                      & 0.6906                                                                      & 0.6962                                                                     & 0.701                                                                     & 0.6883                                                                       \\
en-de                                                         & 0.8113    & 0.7876 & 0.786                                                  & 0.7888                                               & 0.782  & 0.8021      & 0.7911 & 0.7769 & 0.7775                                                                      & 0.7779                                                                      & 0.7805                                                                     & 0.7826                                                                    & 0.7870                                                                       \\
en-pl                                                         & 0.7793    & 0.7442 & 0.738                                                  & 0.7426                                               & 0.7323 & 0.7665      & 0.7491 & 0.7318 & 0.731                                                                       & 0.7303                                                                      & 0.7337                                                                     & 0.7365                                                                    & 0.7429                                                                       \\
en-uk                                                         & 0.7711    & 0.7107 & 0.7193                                                 & 0.7248                                               & 0.7144 & 0.7516      & 0.7199 & 0.7157 & 0.7145                                                                      & 0.7135                                                                      & 0.718                                                                      & 0.7216                                                                    & 0.7246                                                                       \\
pl-de                                                         & 0.7489    & 0.665  & 0.6772                                                 & 0.6861                                               & 0.6652 & 0.6556      & 0.6808 & 0.6841 & 0.6853                                                                      & 0.6861                                                                      & 0.6914                                                                     & 0.6961                                                                    & 0.6852                                                                       \\
pl-en                                                         & 0.7562    & 0.7303 & 0.7343                                                 & 0.7365                                               & 0.7319 & 0.7326      & 0.7345 & 0.73   & 0.7314                                                                      & 0.7325                                                                      & 0.7344                                                                     & 0.7361                                                                    & 0.7351                                                                       \\
pl-uk                                                         & 0.7476    & 0.6776 & 0.6651                                                 & 0.6542                                               & 0.6775 & 0.7031      & 0.6914 & 0.6584 & 0.6647                                                                      & 0.6696                                                                      & 0.6773                                                                     & 0.6838                                                                    & 0.6809                                                                       \\
uk-de                                                         & 0.7242    & 0.6277 & 0.6496                                                 & 0.6603                                               & 0.6347 & 0.6404      & 0.6488 & 0.6595 & 0.6623                                                                      & 0.6644                                                                      & 0.6699                                                                     & 0.6745                                                                    & 0.6597                                                                       \\
uk-en                                                         & 0.7595    & 0.7129 & 0.724                                                  & 0.7281                                               & 0.7185 & 0.7277      & 0.7221 & 0.7231 & 0.7257                                                                      & 0.7278                                                                      & 0.7305                                                                     & 0.7329                                                                    & 0.7277                                                                       \\
uk-pl                                                         & 0.7245    & 0.6629 & 0.6458                                                 & 0.6416                                               & 0.6546 & 0.6845      & 0.6746 & 0.6455 & 0.6511                                                                      & 0.6555                                                                      & 0.6622                                                                     & 0.6679                                                                    & 0.6642                                                                       \\ \hline
alphanumeric                                                  & 0.7507    & 0.699  & 0.7037                                                 & 0.707                                                & 0.7005 & 0.7079      & 0.7089 & 0.7085 & 0.7093                                                                      & 0.71                                                                        & 0.7134                                                                     & 0.7164                                                                    & 0.7113                                                                       \\
email                                                         & 0.7338    & 0.6719 & 0.6821                                                 & 0.6869                                               & 0.6761 & 0.6925      & 0.6847 & 0.6883 & 0.688                                                                       & 0.6878                                                                      & 0.6918                                                                     & 0.6952                                                                    & 0.6899                                                                       \\
emoji                                                         & 0.7606    & 0.7045 & 0.7041                                                 & 0.7079                                               & 0.6989 & 0.7207      & 0.7121 & 0.7069 & 0.709                                                                       & 0.7107                                                                      & 0.715                                                                      & 0.7186                                                                    & 0.7141                                                                       \\
iban                                                          & 0.7616    & 0.7043 & 0.7089                                                 & 0.709                                                & 0.7072 & 0.7194      & 0.7154 & 0.7113 & 0.7125                                                                      & 0.7134                                                                      & 0.7176                                                                     & 0.7211                                                                    & 0.7168                                                                       \\
ip                                                            & 0.7777    & 0.7278 & 0.7321                                                 & 0.735                                                & 0.7278 & 0.7382      & 0.7378 & 0.7357 & 0.7368                                                                      & 0.7375                                                                      & 0.741                                                                      & 0.7439                                                                    & 0.7393                                                                       \\
isbn                                                          & 0.772     & 0.7178 & 0.7238                                                 & 0.7276                                               & 0.7205 & 0.7291      & 0.7292 & 0.7286 & 0.7295                                                                      & 0.7301                                                                      & 0.7338                                                                     & 0.7369                                                                    & 0.7316                                                                       \\
phone                                                         & 0.766     & 0.7127 & 0.7147                                                 & 0.7175                                               & 0.7119 & 0.7236      & 0.7228 & 0.7183 & 0.7193                                                                      & 0.72                                                                        & 0.7237                                                                     & 0.7269                                                                    & 0.7231                                                                       \\
social                                                        & 0.769     & 0.7082 & 0.7171                                                 & 0.722                                                & 0.7113 & 0.7285      & 0.7194 & 0.7215 & 0.7229                                                                      & 0.7239                                                                      & 0.7279                                                                     & 0.7313                                                                    & 0.7253                                                                       \\
url                                                           & 0.7471    & 0.6849 & 0.6963                                                 & 0.7005                                               & 0.6903 & 0.701       & 0.6973 & 0.7014 & 0.7025                                                                      & 0.7034                                                                      & 0.7074                                                                     & 0.7107                                                                    & 0.7036                                                                       \\ \hline
\begin{tabular}[c]{@{}l@{}}macro avg.\\ by model\end{tabular} & 0.7598    & 0.7035 & 0.7092                                                 & 0.7126                                               & 0.7049 & 0.7179      & 0.7142 & 0.7103 & 0.7117                                                                      & 0.7127                                                                      & 0.7168                                                                     & 0.7202                                                                    &                                                                              \\ \hline
\end{tabular}
}

\label{tab:kiwi}
\end{table*}
\label{app:kiwi}
Table~\ref{tab:kiwi} summarizes translation performance across models, language pairs, and input types. Google Translate leads with a macro-average score of 0.7598, performing consistently well in both high-resource pairs (e.g., de-en: 0.795) and low-resource pairs (e.g., uk-pl: 0.7245). EuroLLM models, particularly the 9B version with focused prompts, achieve strong results with a macro-average of 0.7202, making them the closest competitors. Performance generally declines for low-resource pairs and some specialized input types like email and url. Overall, larger models and prompt optimization (as seen with EuroLLM) significantly enhance performance. It is important to note that this test set is synthetic, and the evaluation process is intended primarily as a sanity check to assess machine translation quality.

\section{Detailed Figures and data analysis}

\begin{figure*}[h!]
    \centering
    \includegraphics[width=0.95\textwidth]{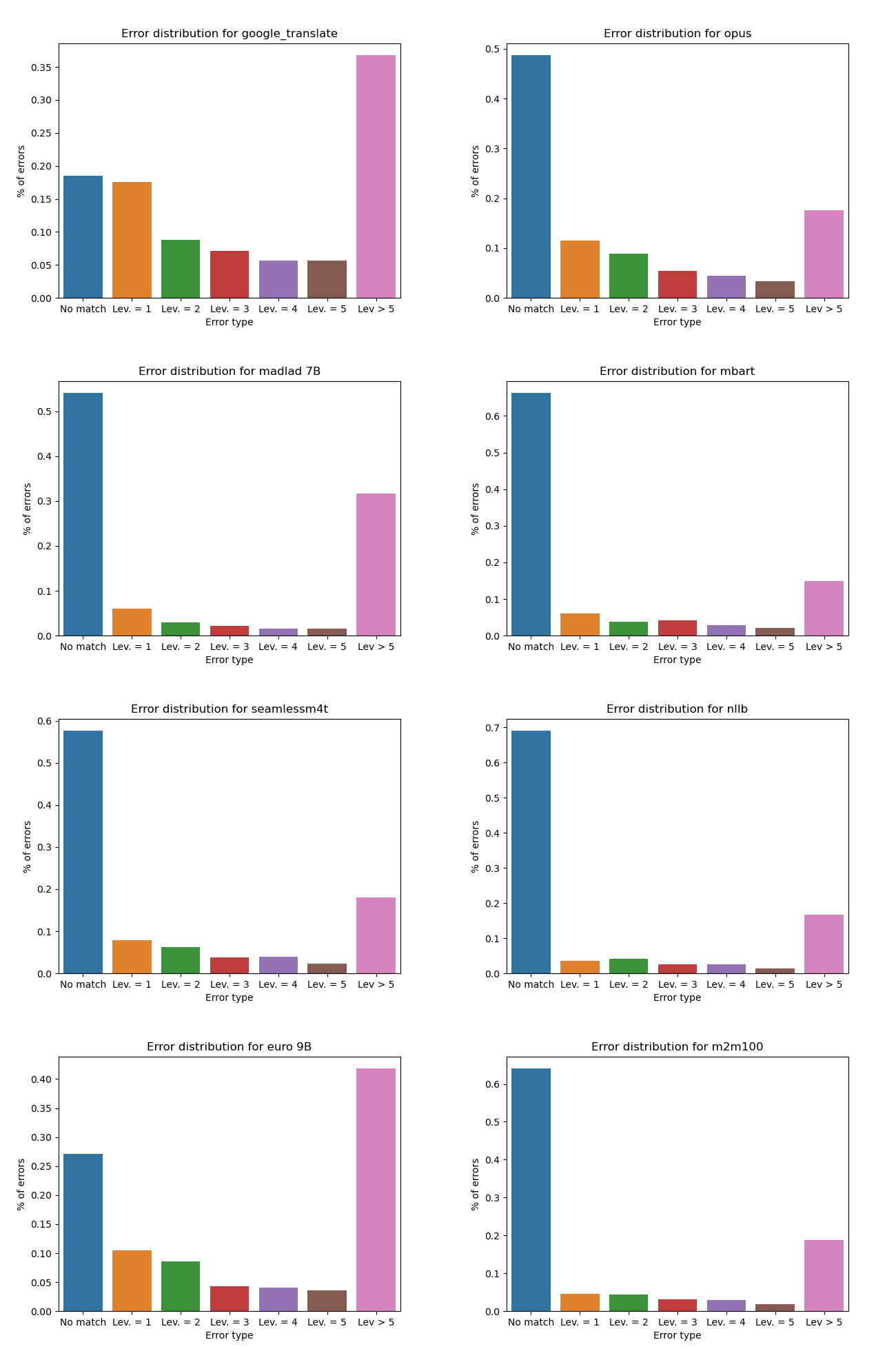}
    \caption{Distribution of the size of errors among models}
    \label{fig:errors}
\end{figure*}

\label{appendix:figures}
\begin{figure*}[h]
    \centering
    \includegraphics[width=1.0\textwidth]{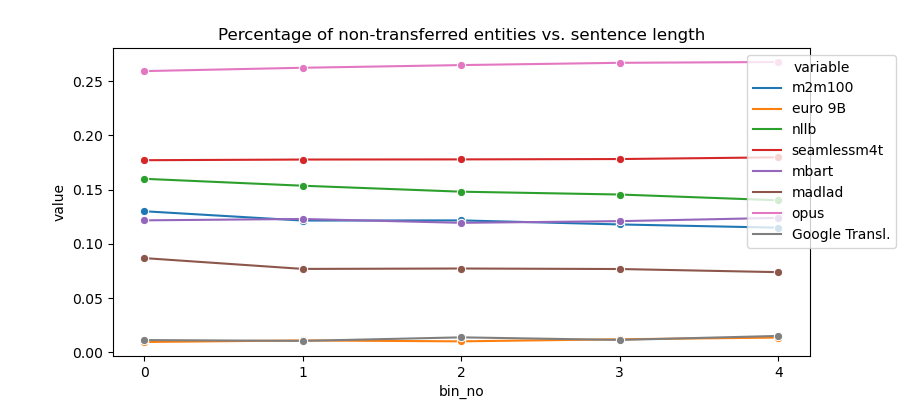}
    \caption{Percentage of entities not matched in target sentences among different lengths.}
    \label{fig:bins_nomatch}
\end{figure*}

\begin{figure*}[h]
    \centering
    \includegraphics[width=1.0\textwidth]{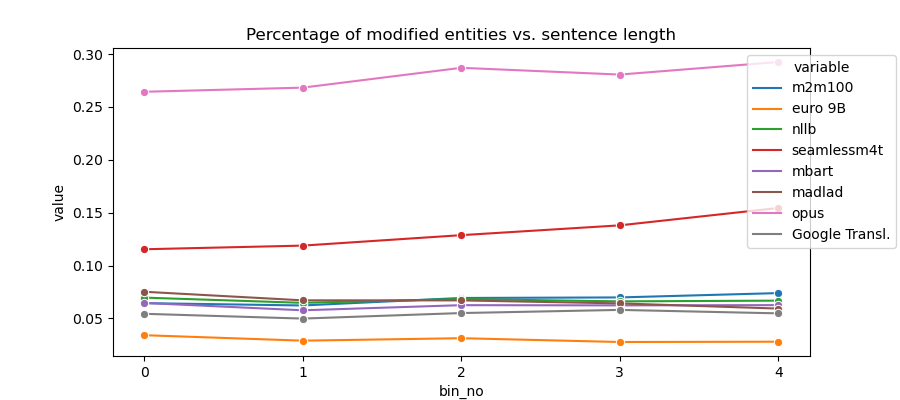}
    \caption{Percentage of modified entities among different lengths.}
    \label{fig:bins_modified}
\end{figure*}

\begin{figure*}[h!]
    \centering
    \includegraphics[width=1\textwidth]{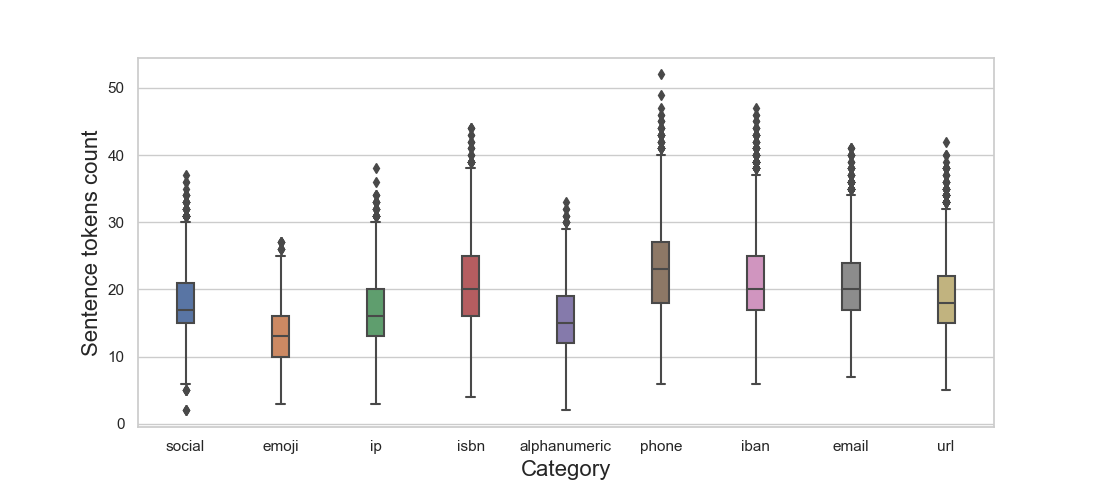}
    \caption{Number of NLTK~\cite{bird2009natural} tokens per category. Sentences expressed in all languages were collected together for this analysis.}
    \label{fig:category_tokens_sentences}
\end{figure*}


\begin{figure*}[h!]
    \centering
    \includegraphics[width=0.95\textwidth]{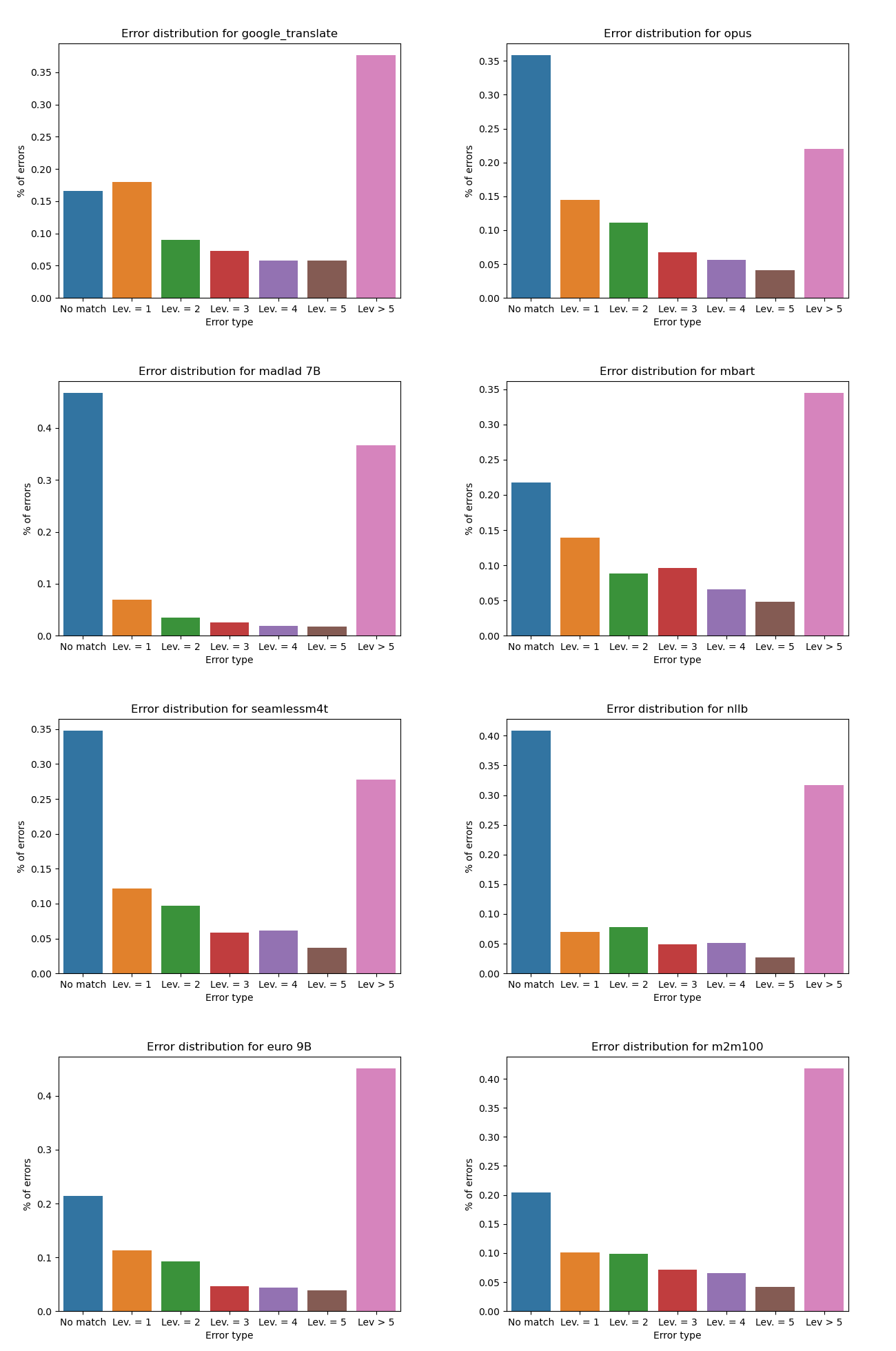}
    \caption{Errors distributions without emoji category}
    \label{fig:errors_dist_no_emoji}
\end{figure*}

\begin{table*}[]

\centering
\scriptsize
\caption{Examples of errors made by Google Translate (verified on 01.01.2025)}
\label{tab:google_translate_errors}
\begin{tabular}{|r|p{14cm}|}
\hline 
comment & \textbf{IBAN modified}, pl $\rightarrow$ de \\
srcEntity & \textbf{PL60109010000000000000000000} \\ 
tgtEntity & \textbf{PL601090100000000000000000} \\ 
srcText & Panie profesorze Janie Kowalski, może przesłać mi nową listę prac do sprawdzenia na konto o numerze PL60109010000000000000000000? \\ 
tgtText & Professor Jan Kowalski, können Sie mir eine neue Liste der zu prüfenden Werke an die Kontonummer PL601090100000000000000000 senden? \\ \hline 
comment & \textbf{Translated social handler}, en $\rightarrow$ de \\
srcEntity & \textbf{@klimatyzacja} \\
tgtEntity & \textbf{@airconditioning} \\
srcText & Przechodząc przez park, nagle usłyszałam @klimatyzacja śpiewającą piosenkę o letnim słońcu. \\
tgtText & Als ich durch den Park spazierte,
hörte ich plötzlich @airconditioning ein
Lied über die Sommersonne singen. \\ \hline 


comment & \textbf{Changed characters in alphanumeric}, en $\rightarrow$ de \\ 
srcEntity & \textbf{tenotypic123CBSprk} \\
tgtEntity & \textbf{tenotypisch123CBSprk} \\ 
srcText & Gniazdo pająka, o symbolu tenotypic123CBSprk, wisiało pod niebem usianym szumami. \\
tgtText & Das Spinnennest, Symbol tenotypisch123CBSprk, hing unter einem mit Lärm übersäten Himmel. \\ \hline 
comment & \textbf{Changed characters in URL}, de $\rightarrow$ en \\ 
srcEntity & \textbf{www.irgendwohin.com} \\
tgtEntity & \textbf{www.somewhere.com} \\
srcText & Die Katze las www.irgendwohin.com vor dem Frühstück und kraulte verschmitzt um Aufmerksamkeit. \\
tgtText & The cat read www.somewhere.com before breakfast and playfully scratched for attention. \\ \hline 
comment & \textbf{Modified e-mail} de $\rightarrow$ en \\
srcEntity & \textbf{liebevollchenpinguin@aya.at} \\ 
tgtEntity & \textbf{lovingchenpinguin@aya.at} \\ 
srcText & Die sprechende Mandarine geschickt ein Bild an liebevollchenpinguin@aya.at. \\ 
tgtText & The talking mandarin sent a picture to lovingchenpinguin@aya.at. \\ \hline

comment & \textbf{IP removed}, pl $\rightarrow$ de \\
srcEntity & \textbf{192.168.1.108} \\
tgtEntity & \textbf{NO ENTITY DETECTED} \\
srcText & Wiatr gonił pożółkłe liście, aż mu się zadało z 192.168.1.108 i spróbowało wziąć pod nie chwytem. \\
tgtText & Der Wind verfolgte die vergilbten Blätter, bis er müde wurde und versuchte, sie zu ergreifen. \\ \hline 
comment & \textbf{Dropped Phone number}, de $\rightarrow$ pl \\ 
srcEntity & \textbf{49 030 1234567890} \\
tgtEntity & \textbf{NO ENTITY DETECTED} \\
srcText & Die alten Ratten spielten Karten und diskutierten leidenschaftlich laut vor Telefonnummer +49 030 1234567890 verband 123politisch. \\
tgtText & Stare szczury grały w karty i głośno i namiętnie dyskutowały o polityce. \\ \hline 
comment & \textbf{ISBN dropped} pl $\rightarrow$ de \\ 
srcEntity & \textbf{978-83-12-34567-8} \\
tgtEntity & \textbf{NO ENTITY DETECTED} \\
srcText & Podczas lekcji astronomii, Paweł natknął się na książkę o istocie czasoprzestrzeni, której ISBN 978-83-12-34567-8 zdradził tajemnicę kosmicznej harmonii. \\
tgtText & Während einer Astronomiestunde stieß Paweł auf ein Buch über die Natur der Raumzeit, das das Geheimnis der kosmischen Harmonie enthüllte. \\ \hline 
\end{tabular}
\end{table*}


\end{document}